\pdfoutput=1

\documentclass[11pt]{article}

\usepackage[preprint]{acl}
\usepackage{listings}
\usepackage{xcolor} 
\usepackage{algorithm}
\usepackage{algorithmic}
\usepackage{multirow} 
\usepackage{booktabs}
\usepackage{amsmath}
\usepackage{amssymb}
\usepackage{tabularx}
\lstdefinestyle{promptStyle}{
    basicstyle=\ttfamily\footnotesize, 
    breaklines=true,                    
    frame=single,                       
    backgroundcolor=\color{gray!10},    
    keywordstyle=\color{blue},          
    commentstyle=\color{green!50!black},
    stringstyle=\color{orange},           
    numbers=left,                       
    numberstyle=\tiny\color{gray},      
    stepnumber=1,                       
    tabsize=4,                          
    showstringspaces=false              
}

\usepackage{times}
\usepackage{latexsym}

\usepackage[T1]{fontenc}

\usepackage[utf8]{inputenc}

\usepackage{microtype}

\usepackage{inconsolata}

\usepackage{graphicx}

\NewDocumentCommand{\cheng}{ mO{} }{\textcolor{blue}{\textsuperscript{\textit{Cheng}}\textsf{\textbf{\small[#1]}}}}

\title{ORBIT: Cost-Effective Dataset Curation for Large Language Model Domain Adaptation with an Astronomy Case Study}

\author{
  Eric Modesitt, Ke Yang, Spencer Hulsey, Chengxiang Zhai \\
  University of Illinois at Urbana-Champaign \\[1ex]
  \textbf{Volodymyr Kindratenko} \\
  National Center for Supercomputing Applications
}

\begin{document}
\maketitle
\begin{abstract}
Recent advances in language modeling demonstrate the need for high-quality domain-specific training data, especially for tasks that require specialized knowledge. General-purpose models, while versatile, often lack the depth needed for expert-level tasks because of limited domain-specific information. Domain adaptation training can enhance these models, but it demands substantial, high-quality data. To address this, we propose ORBIT, a cost-efficient methodology for curating massive, high-quality domain-specific datasets from noisy web sources, tailored for training specialist large language models. Using astronomy as a primary case study, we refined the 1.3T-token FineWeb-Edu dataset into a high-quality, 10B-token subset focused on astronomy. Fine-tuning \textsc{LLaMA-3-8B} on a 1B-token astronomy subset improved performance on the MMLU astronomy benchmark from 69\% to 76\% and achieved top results on AstroBench, an astronomy-specific benchmark. Moreover, our model (Orbit-LLaMA) outperformed \textsc{LLaMA-3-8B-base}, with GPT-4o evaluations preferring it in 73\% of cases across 1000 astronomy-specific questions. Additionally, we validated ORBIT's generalizability by applying it to law and medicine, achieving a significant improvement of data quality compared to an unfiltered baseline. We open-source the ORBIT methodology, including the curated datasets, the codebase, and the resulting model at \href{https://github.com/ModeEric/ORBIT-Llama}{https://github.com/ModeEric/ORBIT-Llama}.

\end{abstract}

\section{Introduction}

The rapid advancement of large language models (LLMs) has transformed natural language processing (NLP) and artificial intelligence (AI), with general-purpose models like GPT-4 \citep{openai2024gpt4} and LLaMA \citep{Llama3Team2024} demonstrating versatility across tasks such as knowledge retrieval, open-domain question answering, and linguistic applications. However, these models often struggle in specialized domains, such as astronomy, where deep, nuanced understanding and up-to-date factual accuracy are crucial \cite{singhal_large_2023}. This performance gap arises because general-purpose LLMs must balance performance across a wide range of tasks, diluting domain-specific knowledge \cite{li_revisiting_2024, yang_unveiling_2024}.

To address this limitation, domain-specialized LLMs can allocate their capacity toward mastering specific domains, offering greater depth and accuracy. However, building these models is challenging due to the need for high-quality, domain-specific datasets. Conventional approaches, such as using academic sources like arXiv papers \citep{nguyen_astrollama:_2023,pan_astromlab_2024}, tend to focus on highly technical content, neglecting the breadth and diversity needed for effective model generalization. Alternatively, web-sourced datasets offer greater diversity but are often noisy, containing irrelevant or low-quality content. Traditional filtering methods, such as keyword-based or rule-based approaches, frequently fail to balance coverage and quality, potentially excluding relevant data while admitting suboptimal material.

In this work, we propose \textbf{ORBIT}, a novel, scalable data curation framework for creating high-quality, domain-specific datasets. ORBIT combines embedding-based similarity matching with a BERT-based regression model to filter large-scale web datasets efficiently. By focusing on both semantic relevance and educational value, this methodology ensures that the curated datasets are both diverse and tailored to specific domains. Using astronomy as the primary case study, we curated a 10-billion-token dataset derived from FineWeb-Edu \cite{penedo_fineweb_2024}, incorporating a broader range of content compared to prior approaches like AstroLLaMA \cite{nguyen_astrollama:_2023}, which rely solely on arXiv abstracts. The inclusion of web-sourced educational content alongside academic texts enables ORBIT to balance depth and diversity, capturing a more comprehensive understanding of domain-specific knowledge.

To demonstrate the generalizability of ORBIT, we also applied it to law and medicine, achieving significant quality improvements in these domains. GPT-4o evaluations rated the curated datasets at an average educational value of 3.05 and 2.9 on a scale of 0-5 per document, respectively, compared to an unfiltered baseline of approximately 0.4. These results highlight ORBIT’s ability to extract domain-relevant, high-quality data across diverse fields.

Fine-tuning a \textsc{LLaMA-3-8B} model on a randomly sampled 1B-token astronomy subset of the ORBIT-curated dataset results in substantial improvements on astronomy-specific tasks. Our model (Orbit-LLaMA) achieves a 7-point accuracy gain over the base LLaMA-3-8B model (from 69.08\% to 76.3\%) on the MMLU astronomy benchmark and outperforms AstroLLaMA (66.45\%) by a significant margin. Furthermore, ORBIT-trained models surpass state-of-the-art performance on various astronomy baselines, receiving higher ratings from both GPT-4o evaluations and domain experts in the vast majority of cases. These results underscore the value of ORBIT's methodology in producing specialized datasets that enhance both the depth and breadth of domain-specific knowledge in LLMs.

The key contributions of this paper are:
\begin{itemize}
\item We introduce \textbf{ORBIT}, a generalizable, scalable framework for filtering noisy web data into high-quality, domain-specific datasets, addressing challenges of scalability, noise, and coverage balance.
\item We demonstrate ORBIT's generalizability by applying it to multiple domains, including astronomy, law, and medicine, achieving significant quality improvements in each field with minimal computational overhead.
\item We present a \textbf{specialized astronomy dataset} curated using ORBIT, comprising 10 billion tokens that combine academic rigor with web-scale diversity, advancing prior work limited to arXiv-based sources.
\item We train a \textbf{state-of-the-art astronomy-specific language model} (which we call Orbit), fine-tuned on a subset of the ORBIT-curated dataset, achieving significant performance gains on astronomy-related benchmarks and surpassing existing models, including AstroLLaMA, in expert evaluations.
\end{itemize}

By presenting ORBIT and its application to astronomy, as well as its successful extension to law and medicine, we provide a generalizable framework for developing targeted, domain-specific AI tools. This methodology has the potential to accelerate scientific research, education, and practical applications across a wide range of specialized fields.

\section{Related Work}

\subsection{Data Curation for Language Models}

Recent research has demonstrated the paramount role of high-quality data in the development of large language models. For instance, the technical reports of models like LLama-3 \citep{grattafiori_llama_2024} and Qwen-2 \citep{yang_qwen2_2024} emphasize extensive data curation methodologies for general-purpose language models. These efforts have led to significant performance gains, even when model architectures and parameter sizes remain largely unchanged (e.g., the transition from LLama-2 to LLama-3).

Several efforts have focused on automated data curation techniques. \citet{chen_data-juicer:_2023} proposed a method to automatically filter and clean web-crawled data to build high-quality training corpora, while \citet{gururangan_dont_2020} developed a data selection method for identifying domain-relevant examples within large datasets. Furthermore, \citet{kreutzer_quality_2022} demonstrated that smaller, carefully curated datasets often outperform larger but noisier datasets.

However, these methods often face limitations when applied to highly specialized domains. Many automated filtering techniques rely on general quality metrics or term whitelisting, which can inadvertently include irrelevant or low-quality content while excluding high-quality data that does not fit predefined patterns. For instance, filtering by specific terms or phrases, such as LaTeX commands, may be effective in domains like mathematics but fails in more diverse fields like astronomy where specialized exact terms do not exist or are more varied. Additionally, many datasets rely on scraped web data, which presents risks related to copyright issues, noise, and incomplete data extraction from APIs, further limiting the potential for domain-specific curation.

\subsection{Domain-Specific Language Models}

Advances in natural language processing have led to the rise of domain-specific language models that are fine-tuned on specialized corpora. 
These models are designed to perform well within particular domains, outperforming general-purpose models on domain-specific tasks \citep{ beltagy_scibert:_2019}. However, each of these approaches has notable limitations.

For example, \citet{azerbayev_llemma:_2024} introduced LLEMMA, an open-source language model for mathematics that achieves state-of-the-art results on the MATH benchmark. LLEMMA filters data based on whether it contains LaTeX syntax, a technique well-suited to mathematics but restrictive when applied to other fields, such as astronomy or biology, where such syntactic markers do not exist. This method risks excluding valuable content that lacks LaTeX or including low-quality data simply because it contains LaTeX markup.

Similarly, \citet{singhal_large_2023} developed Med-PaLM 2, a medical domain model that achieved 85.4\% accuracy on US Medical Licensing Examination (USMLE) questions. However, its approach to fine-tuning is relatively limited, relying primarily on instruction fine-tuning without deep post-training adjustments specific to medical literature, limiting its adaptability for more niche medical tasks.

Other domain-specific models face similar limitations in data sourcing. \citet{yang_fingpt:_2023} introduced FinGPT, which demonstrates strong performance on financial tasks, but it heavily relies on domain-specific data sources like SEC filings and NYSE transaction reports. These data sources are highly specific to the financial domain and do not generalize well to other fields, limiting the flexibility of such models.

\citet{nguyen_astrollama:_2023} introduced AstroLLaMA, a 7-billion-parameter model fine-tuned on the abstracts of 300,000 astronomy papers from arXiv. Furthermore, \citet{ting_astromlab_2024} builds upon this work with larger and more modern models. While these works show strong performance in generating scientifically relevant text completions, limiting the dataset to only arXiv papers (and in this case, only to certain sections such as the Abstract and Introduction) restricts the breadth and depth of the information available for fine-tuning. The homogeneous distribution of similarly formatted research abstracts leads to a lack of data diversity that reduces the model's capacity to generalize across broader applications within the domain.

These models highlight the importance of high-quality, domain-specific datasets for effective model performance but also demonstrate the challenges in collecting and curating sufficiently diverse and representative datasets.
\section{Dataset Curation Methodology}

\subsection{Choice of Corpus}

\begin{algorithm}[tb]
   \caption{Orbit Domain-Specific Dataset Curation Pipeline}
   \label{alg:dataset_curation}
\begin{algorithmic}
   \STATE {\bfseries Input:} Corpus of documents, astronomy-related terms, similarity threshold $\tau$, educational value threshold $\eta$
   \STATE {\bfseries Output:} Filtered astronomy-specific dataset
   \STATE Initialize astronomy vector $\mathbf{A}$ by averaging embeddings of astronomy-related terms
   \STATE {\bfseries Stage 1: Embedding-Based Threshold Filtering}
   \FOR{each document $D$ in the corpus}
       \STATE Compute document vector $\mathbf{B}$ by aggregating embeddings of tokens in $D$
       \STATE Calculate similarity: $\text{Similarity}(D) = \frac{\mathbf{A} \cdot \mathbf{B}}{|A|*|B|}$
       \IF{$\text{Similarity}(D) > \tau$}
           \STATE Retain document $D$
       \ENDIF
   \ENDFOR
   \STATE {\bfseries Stage 2: BERT-Based Regressor Evaluation}
   \FOR{each retained document $D$}
       \STATE Compute educational value score $EV(D)$ using BERT-based regressor
       \IF{$EV(D) > \eta$}
           \STATE Retain document $D$
       \ENDIF
   \ENDFOR
   \STATE {\bfseries Return} filtered dataset
\end{algorithmic}
\end{algorithm}
\label{sec:dataset_curation_methodology}

\begin{figure*}[htbp]
    \centering
    \includegraphics[width=0.7\textwidth]{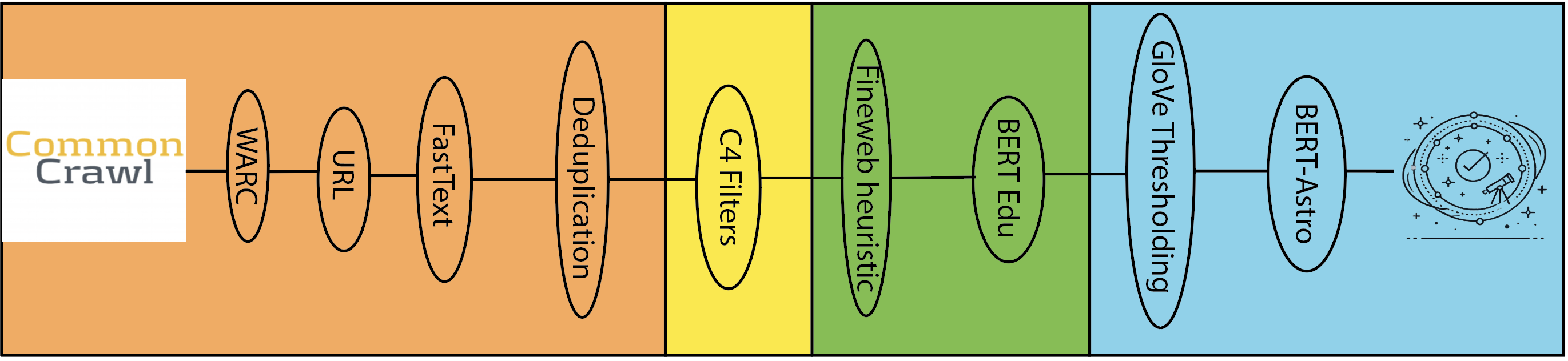}
    \caption{Comprehensive Filtering Pipeline from FineWeb-Edu to ORBIT. The pipeline emphasizes the quality and size of the dataset. The orange includes common filtering methods formalized in \citet{wenzek_ccnet:_2020}. The yellow summarizes large-scale semantic filters from \citet{raffel_exploring_2023}. The green includes the additional semantic filters and the BERT-based classifier used to filter for educational relevance in FineWeb-Edu. The blue outlines our contributions: GloVe-based embedding thresholding and a BERT classifier for educational relevance specific to astronomy. See subsections \ref{sec:dataset_curation} and \ref{sec:educational_assessment} for details on our contributions.}
    \label{fig:sample_image}
\end{figure*}

For this study, we selected the FineWeb-Edu dataset \cite{penedo_fineweb_2024} as our primary corpus. FineWeb-Edu is a specialized subset of FineWeb, which is a large-scale, high-quality dataset derived from CommonCrawl web data, specifically designed for pretraining large language models. The FineWeb-Edu dataset uses the Open Data Commons License Attribution family. FineWeb-Edu focuses on ``educational content'' based on prompt engineering strategies and contains approximately 1.3 trillion tokens, curated by filtering out content with lower educational value. This subset allowed us to begin with a high-quality dataset that is more focused and manageable for the specific tasks required in astronomy. Figure \ref{fig:sample_image} illustrates the comprehensive filtering pipeline from FineWeb-Edu to ORBIT (our method), highlighting the quality and size at each step with examples of what has been eliminated.

\subsection{Methodology for Domain-Specific Dataset Curation}
\label{sec:dataset_curation}

Our research presents a novel approach to curating a high-quality, domain-specific dataset for astronomical language models. This methodology combines advanced natural language processing techniques with rigorous quality assurance measures to produce a dataset that balances complex reasoning tasks with factual content in the field of astronomy. Our approach is designed for cost-effectiveness, using a combination of broad initial filtering and more thorough assessments at later stages to optimize the dataset's quality and relevance. Our full filtering method is shown in Algorithm \ref{alg:dataset_curation}.

\subsubsection{Stage 1: Initial Domain-Specific Filtering}

We developed a lexicon of 101 single-word astronomy-related terms, encompassing concepts from astrophysics, cosmology, and space exploration. To efficiently process large volumes of text, we implemented a static-embedding-based matching technique utilizing GloVe word embeddings \citep{pennington_glove:_2014}. A representative astronomy aggregated embedding vector \( \mathbf{A} \) was computed by averaging the embeddings of all terms. For each document in FineWeb-Edu, we calculated a document vector and computed its cosine similarity with \( \mathbf{A} \). Documents exceeding a similarity threshold of $\tau=0.2$ were retained for further analysis. This threshold was empirically determined to balance dataset size and quality. After this stage, approximately 20B tokens of the corpus remained.

\subsubsection{Stage 2: Educational Value Assessment}
\label{sec:educational_assessment}

\begin{figure}[ht]
    \centering
\includegraphics[width=0.48\textwidth]{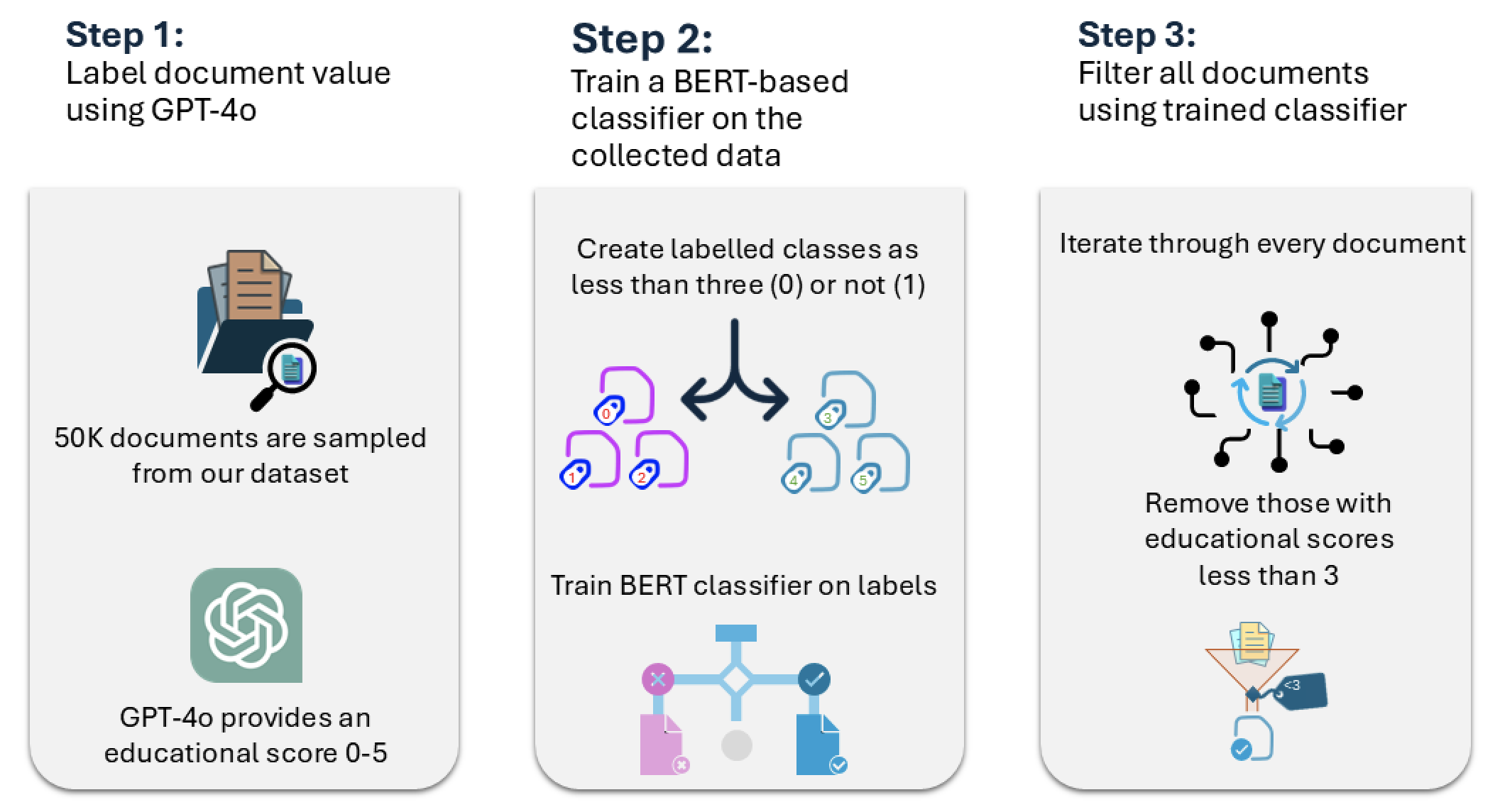}
    \caption{
        Full Stage 2 pipeline visualized. 
    }
    \label{fig:bert_classification}
\end{figure}
\begin{figure}[ht]
    \centering
    \includegraphics[scale=0.24]{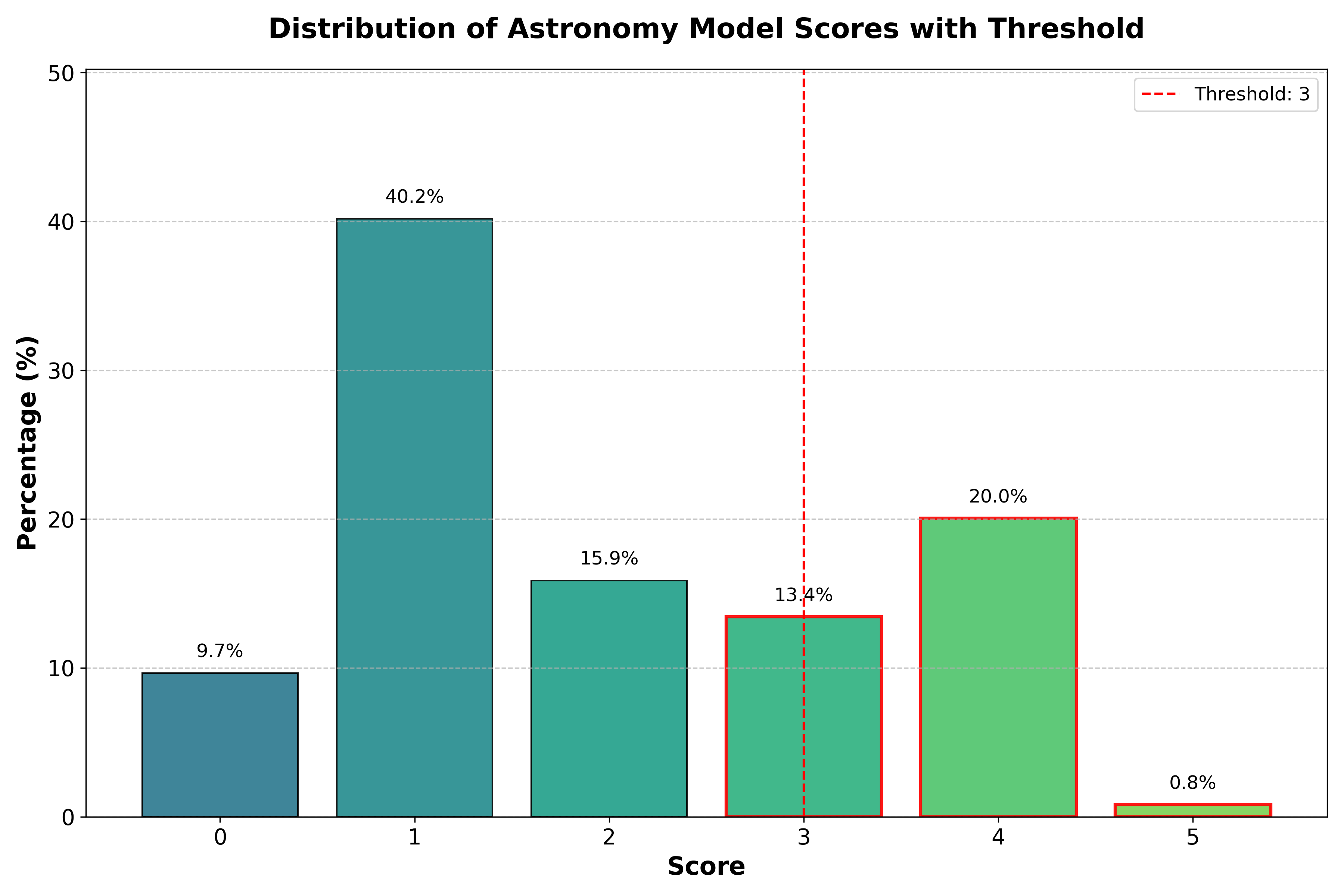}
    \caption{ Distribution of educational value scores (ranging from 0 to 5) assigned by the BERT-based regressor model to a sample of 1000 astronomy-related documents. This visualization demonstrates the validity of the classifier by showing alignment with expected distributions based on held-out test sets and expert evaluations.}
    \label{fig:score_distribution}
\end{figure}

After the initial filtering, we applied a more thorough evaluation to refine the dataset further, focusing on its educational merit. Without this second phase, we would be left with a number of low-qualtiy documents, as shown in  Figure \ref{fig:score_distribution}. Furthremore, if only Stage 2 was applied, the computational cost would increase significantly. For example, if Stage 1 keeps $\frac{1}{100}$ of the total data, the number of NVIDIA A100 GPU hours needed for stage 2 would decrease by 100x. See Table \ref{tab:filtering_comparison} for more information.
\begin{table*}[h!]
\centering
\small 
\caption{Comparison of Processing Time and Cost for Dataset Filtering. Stage 1 filtering retains 1\% of documents (and thus tokens), drastically reducing the effective dataset size for Stage 2. Stage 2 alone processes the full dataset. The combined approach significantly lowers the time and cost of Stage 2. Pricing estimates are based on current market rates and hardware usage. Furthermore, both stages are fully parallelizable, meaning additional hardware can cause linear decrease in time for an approximately constant price.}
\begin{tabular}{l|lcll}
\toprule
\textbf{Scenario}    & \textbf{Processing Unit}         & \textbf{Total Time}           & \textbf{Total Cost}                       & \textbf{Quality} \\ \midrule
Stage 1 Only         & Intel Core i9 (16 cores)         & 177 hours                  & \$44                                   & Medium           \\ 
Stage 2 Only         & A100 PCIe GPU (1 unit)          & 12,000 hours                  & \$16,200                               & Highest          \\ 
Stage 1 + Stage 2    & Intel Core i9 + A100 PCIe GPU    & 297 hours  & \$206                                  & Highest          \\ \bottomrule
\end{tabular}
\label{tab:filtering_comparison}
\end{table*}

We developed a BERT-based regressor model \citep{devlin_bert:_2019}, using Huggingface's \textsc{HuggingFaceFW/fineweb-edu-classifier} model, trained to evaluate the educational value of astronomy-related text on a scale of 0 to 5.

The training dataset for this model was meticulously curated through a multi-step process:
\begin{enumerate}
    \item Random sampling of 50,000 documents from the embedding-filtered corpus to ensure topic diversity.
    \item Automated evaluation of each sampled document using GPT-4o model \citep{openai_gpt-4o_2024}, which was prompted to assess the educational value on a 6-point scale (0-5).
    \item Collection of both quantitative scores and qualitative justifications for each evaluation, used for prompt engineering.
\end{enumerate}

The language model was instructed to consider factors such as depth of astronomical content, clarity of explanations, relevance to a general audience, and the presence of advanced concepts. Our prompt, inspired by \citet{yuan_self-rewarding_2024} (see Appendix), emphasized educational value specific to the domain of astronomy. See Figure \ref{fig:bert_classification} for a visual of our Stage 2 pipeline. We kept any value above or equal to our threshold $\eta=3$, resulting in approximately 10 billion tokens of high-quality, astronomy-relevant content.

\subsubsection{Cross-Domain Validation: Law and Medicine}

To assess the generalizability of ORBIT, we extended the dataset curation pipeline to two additional domains: law and medicine. Using the same methodology applied to astronomy, we developed domain-specific lexicons for these fields. For law, the lexicon included terms such as “litigation,” “precedent,” and “contract,” while for medicine, it featured terms like “pathology,” “oncology,” and “metastasis.” The complete lists of terms for each domain are provided in the Appendix.

Stage 1 filtering, based on embedding-based similarity, was adapted to these domains by computing aggregated embedding vectors from their respective lexicons. For each document in FineWeb-Edu, the cosine similarity between its embedding vector and the domain-specific aggregated vector was calculated. Documents exceeding the similarity threshold of 0.2 were retained for further analysis.

\section{Experiments}

To validate the effectiveness of the ORBIT methodology, we conducted a series of experiments focusing on the quality of the curated dataset, the impact of fine-tuning on model performance, and the influence of different thresholding values within the pipeline. These experiments aim to assess how ORBIT's two-stage filtering approach improves dataset relevance and educational value while balancing dataset size and computational cost. Additionally, we evaluate the performance of models fine-tuned on ORBIT-curated datasets with varying similarity and educational value thresholds, examining their impact on downstream tasks. The results provide insights into the trade-offs between dataset size, quality, and curation efficiency, while demonstrating the effectiveness of ORBIT for training astronomy-specialized language models. Below, we outline the experimental setup, datasets, and evaluation metrics used to address these questions.

\subsection{Experimental Setup}

For our experiments, we utilized the Delta GPU cluster at the National Center for Supercomputing Applications, equipped with 8 NVIDIA A100 GPUs, each with 40GB of memory. The model, named Orbit-LLaMA, was derived from Meta's LLaMA-3-8B \citep{Llama3Team2024}, an 8-billion-parameter language model optimized for large-scale training. We used the Punkt tokenizer from NLTK for sentence segmentation during preprocessing. LLaMA-3 operates under the LLaMA 3 Community License Agreement. See the Appendix for more training details.

\subsection{Effect of Thresholding, Embedding Methods, and Keyword Search}

To explore the effectiveness of various filtering strategies, we tested the impact of:
\begin{enumerate}
    \item Different threshold values in embedding similarity filters.
    \item Multiple embedding methods, including fastText, 100-dimensional, and 300-dimensional embeddings.
    \item Keyword filtering approaches compared to unfiltered datasets.
\end{enumerate}

\begin{figure}[h]
    \centering
    \includegraphics[width=\linewidth]{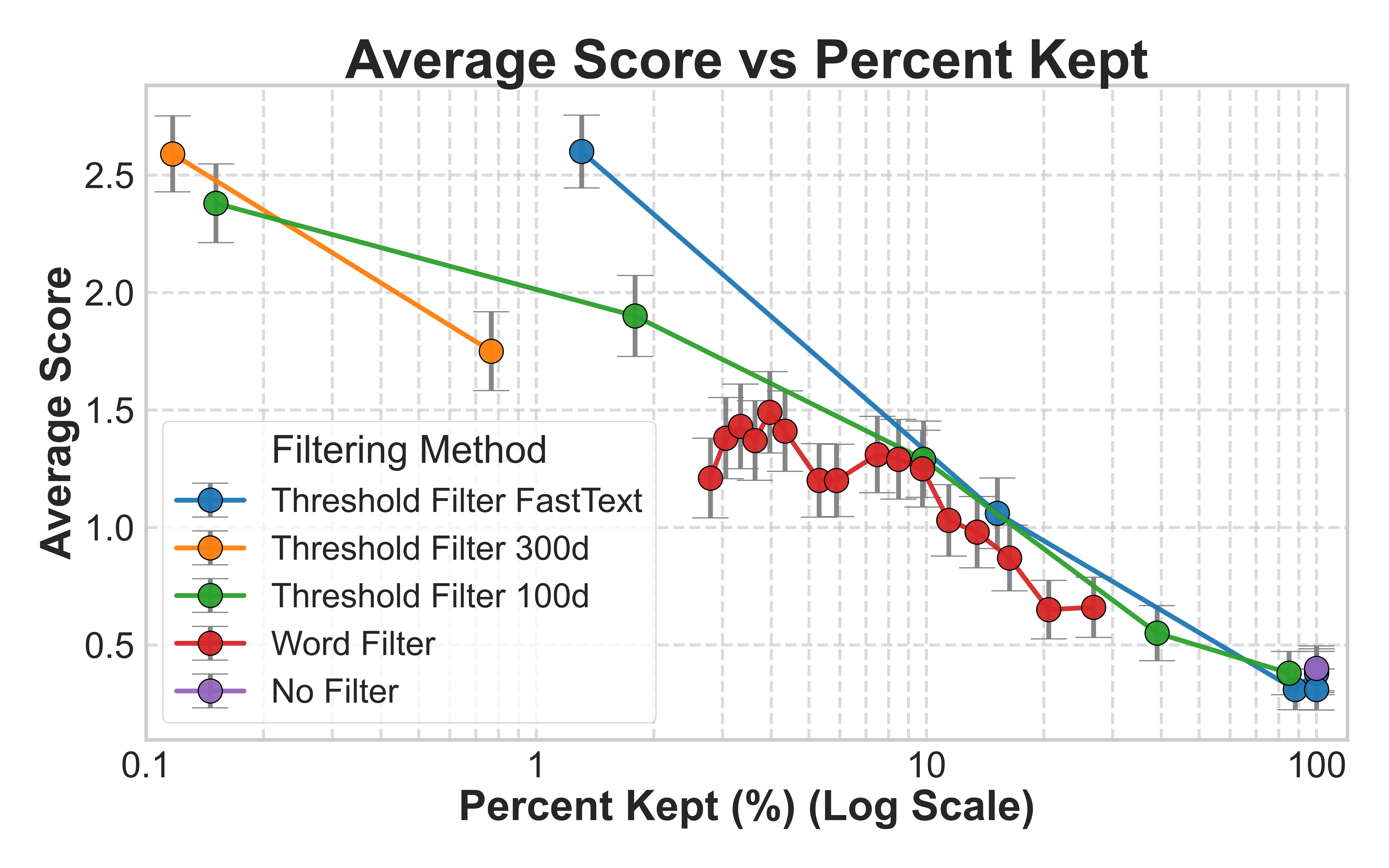}
    \caption{Average Score vs Percent Kept, comparing different filtering methods: embedding thresholds (fastText, 100d, 300d), keyword filtering, and no filtering. The x-axis is log-scaled for clarity.}
    \label{fig:thresholding_experiment}
\end{figure}

This analysis assessed how these methods balance dataset quality and coverage. The performance of each filtering strategy was measured based on average scores obtained from downstream tasks, as shown in Figure~\ref{fig:thresholding_experiment}. Error bars indicate the standard error of the mean (SEM), highlighting variability. The results underscore how keyword filters and embedding-based thresholds can improve dataset curation by focusing on the most relevant content.

The results demonstrate that:
\begin{itemize}
    \item Higher threshold values generally reduce dataset size while maintaining or improving average scores.
    \item Embedding methods showed slightly varying efficacy.
    \item Keyword filtering, while simpler, achieved competitive performance by focusing on domain-specific terminology.
    \item No filtering resulted in the largest datasets but the lowest scores.
\end{itemize}

\subsection{Cross-Domain Validation: Law and Medicine}

To evaluate ORBIT’s generalizability, we applied the dataset curation pipeline to two additional domains: law and medicine. Stage 1 filtering was adapted to these domains by constructing domain-specific lexicons, following the methodology described in Section~\ref{sec:dataset_curation}. For law, the lexicon included terms such as “litigation,” “precedent,” and “contract,” while for medicine, it featured terms like “pathology,” “oncology,” and “metastasis” (see Appendix for full term lists). 

Embedding-based similarity filtering retained approximately 1.0\% of the initial corpus for law and 1.0\% for medicine, similar to the retention rate observed for astronomy. The average educational value scores, evaluated using GPT-4o, showed significant improvements over the unfiltered baseline (0.3), with 2.9 for medicine and 3.05 for law.

These scores align closely with the results obtained for astronomy, indicating that Stage 1 filtering alone is sufficient to extract high-quality, domain-specific content across diverse fields.

\subsection{Benchmarks and Baselines}

We used multiple-choice perplexity prediction to select answers and conducted qualitative pairwise comparisons rated by expert astronomers and GPT-4 for accuracy, clarity, and reasoning. Baseline models were AstroLLaMA-3-8B \citep{pan_astromlab_2024}, the prior state-of-the-art in astronomy language modeling, and Meta-LLaMA-3-8B, a general-purpose model.
\subsubsection{Quantitative Evaluation}
We evaluated Orbit-LLaMa using multiple datasets, including the astronomy section of the MMLU benchmark \citep{hendrycks_measuring_2021} and two versions of AstroBench, the official, validated multiple choice  one\citep{ting_astromlab_2024}, and the Huggingface-only dataset containing three subsets covering important subtests in astronomy \citep{astrobench_basic_knowledge_test,astrobench_knowledge_application,astrobench_scientific_calculation_test}.

A total of three datasets were used for quantitative analysis:

\begin{enumerate}
    \item \textbf{MMLU Benchmark}: The astronomy section of MMLU evaluates factual knowledge and reasoning across topics like stellar formation and cosmology, testing scientific depth in language models.
    \item \textbf{Hugging Face AstroBench Subcategories}: Organized into subcategories:
        \begin{itemize}
            \item \textbf{Basic Knowledge (BK)}: Tests core astronomy concepts.
            \item \textbf{Scientific Calculation (SC)}: Involves solving astrophysical numerical problems.
            \item \textbf{Knowledge Application (KA)}: Assesses applying knowledge to novel scenarios.
        \end{itemize}
        Each subcategory is scored separately for detailed performance analysis.

    \item \textbf{Official AstroBench Benchmark}: A comprehensive dataset of 4,425 multiple-choice questions from 885 \textit{Annual Review of Astronomy and Astrophysics} articles (1963–2023). It provides an aggregated performance score, covering diverse topics such as quasars, cosmological simulations, and the circumgalactic medium.
\end{enumerate}

\subsubsection{Qualitative Evaluation}
We compared responses from Orbit-LLaMA, AstroLLaMA, and Meta-LLaMA using 24 test questions developed by Astronomy Ph.D. students and faculty. Responses were ranked for accuracy (or, for active areas of research, likelihood), clarity, and reasoning using preference ratings for each model, and detailed feedback on the model's strength's and weaknesses.


\subsection{Experiment Results}

Orbit-LLaMa outperformed baselines on all metrics. On the MMLU astronomy section, Orbit-LLaMa scored 76 compared to 69 (Meta-LLaMA) and 66.45 (AstroLLaMA). On AstroBench subcategories, Orbit-LLaMa excelled in Basic Knowledge (45.53\%), Scientific Calculation (30.28\%), and Knowledge Application (45.53\%). On the official AstroBench, Orbit-LLaMa scored 69.7, surpassing AstroLLaMA (66.4) and Meta-LLaMA (61.5).

Table~\ref{tab:shortened_model_comparison} summarizes the results, showing Orbit's superior performance in both specific tasks and overall benchmarks.
\begin{table*}[ht]
    \centering
    \setlength{\tabcolsep}{3pt}
    \renewcommand{\arraystretch}{1.2}
    \caption{Performance Comparison of Models on MMLU and AstroBench Tasks. 
    MMLU sections with problem counts include AS: Astronomy (152), CC: College Chemistry (100), CP: College Physics (102), CPH: Conceptual Physics (235), HSC: High School Chemistry (203), HSP: High School Physics (151). 
    AstroBench sections with problem counts include KA: Knowledge Application (276), SC: Scientific Calculation (251), BK: Basic Knowledge (8772), Astrobench (4425).}
    \label{tab:shortened_model_comparison}
    {\scriptsize
    \begin{tabular}{l|cccccc|ccc|c}
        \toprule
        \textbf{Model} & \textbf{AS (152)} & \textbf{CC (100)} & \textbf{CP (102)} & \textbf{CPH (235)} & \textbf{HSC (203)} & \textbf{HSP (151)} & \textbf{KA (276)} & \textbf{SC (251)} & \textbf{BK (8772)} & \textbf{AstroBench} \\ 
        \midrule
        AstroLLaMA & 66.45 & 47.00 & 38.24 & 55.74 & 53.20 & 41.06 & 39.84 & 29.48 & 63.75 & 66.4 \\ 
        Meta-LLaMA & 69.08 & 44.00 & 37.25 & 54.04 & 52.22 & 41.72 & 41.46 & 25.90 & 65.50 & 61.5 \\ 
        Orbit-LLaMA & \textbf{76.30} & \textbf{52.00} & \textbf{47.10} & \textbf{56.20} & \textbf{53.70} & \textbf{43.10} & \textbf{45.53} & \textbf{30.28} & \textbf{69.96} & \textbf{69.7} \\ 
        \bottomrule
    \end{tabular}}
\end{table*}

Pairwise comparisons confirmed Orbit-LLaMA’s superiority, with win rates over 92\% against baselines (Table~\ref{tab:win_rates_full}). Expert feedback highlighted its accuracy, clarity, and reasoning improvements. See the appendix for detailed examples.

\begin{table*}[ht]
    \centering
    \setlength{\tabcolsep}{3pt}
    \renewcommand{\arraystretch}{1.2}
    \caption{Win Rates and Tie Percentages Between Models.}
    \label{tab:win_rates_full}
    {\scriptsize
    \begin{tabular}{l|cccc}
        \toprule
        \textbf{Models Compared} & \textbf{Meta-LLaMA} & \textbf{Orbit-LLaMA} & \textbf{AstroLLaMA} & \textbf{Tie} \\ \midrule
        Meta-LLaMA vs Orbit-LLaMA & 25.4 & \textbf{73.0} & - & 1.6 \\
        Meta-LLaMA vs AstroLLaMA & \textbf{84.3} & - & 10.5 & 5.22 \\
        Orbit-LLaMA vs AstroLLaMA & - & \textbf{93.0} & 5.0 & 2.0 \\
        \bottomrule
    \end{tabular}}
\end{table*}

Qualitative results by astronomy graduate students further validate these conclusions
\begin{enumerate}
    \item \textbf{Preference Ratings}: Four graduate students selected the best response for each question. Majority consensus was reached for 83\% of questions, with Orbit-LLaMA preferred for 66\% of total responses (Table~\ref{tab:wins-votes}).

\begin{table}[h!]
\centering
\begin{tabular}{lc}
\toprule
\textbf{Model} & \textbf{Selected Output(\%)} \\
\midrule
Meta-LLaMA & 22.1\% \\
Orbit-LLaMA & 66.3\% \\
AstroLLaMA & 11.6\% \\
\bottomrule

\end{tabular}
\caption{The total number of times each model's response was selected from total votes cast (N = 95).}
\label{tab:wins-votes}
\end{table}

    \item \textbf{Detailed Feedback}: Reviewers noted:
        \begin{itemize}
            \item \textbf{Meta-LLaMA}: Responses often repeated content and lacked focus.
            \item \textbf{Orbit-LLaMA}: Delivered clear and concise answers resembling student-created work.
            \item \textbf{AstroLLaMA}: Long, research-style responses with structural and coherence issues.
        \end{itemize}
\end{enumerate}
\section{Discussion}

The results demonstrate the utility of the ORBIT methodology in addressing key challenges in domain-specific dataset curation and fine-tuning. By using a two-stage filtering process, ORBIT balances relevance and quality while remaining computationally efficient. Stage 1’s embedding-based similarity filtering significantly reduces the dataset size, while Stage 2’s educational value assessment ensures the retained data is highly relevant and informative. This layered approach enables the creation of datasets that are both comprehensive and focused, as evidenced by its application to astronomy, law, and medicine.

Fine-tuning Orbit-LLaMA on the ORBIT-curated dataset led to notable improvements across multiple benchmarks, including MMLU astronomy and AstroBench. The gains in both quantitative metrics and qualitative evaluations highlight the impact of curating diverse and high-quality domain-specific data. The inclusion of a mix of academic and educational content allowed the model to excel in tasks requiring both factual knowledge and nuanced reasoning, demonstrating the value of combining depth with breadth in training corpora.

The success of ORBIT in multiple domains also suggests its scalability and adaptability. However, differences in domain-specific challenges, such as interdisciplinary overlaps or evolving knowledge in fields like medicine, highlight the need for further refinement. Future work could focus on automating lexicon creation and optimizing threshold selection to streamline application to new domains.

Overall, the experiments validate the potential of domain-adapted LLMs when supported by robust curation pipelines like ORBIT. This approach addresses limitations in general-purpose models for specialized tasks, emphasizing the importance of targeted datasets for achieving state-of-the-art performance in specific fields.

\section{Conclusion}

This paper presents a novel approach to creating high-quality, domain-specific datasets for training language models, with a focus on the field of astronomy. Our methodology, combining embedding-based matching and BERT-based regression for data filtering and selection, has demonstrated significant potential for enhancing the performance of language models in specialized scientific domains. Furthermore, we validated the scalability and generalizability of this approach by extending it to the domains of law and medicine, achieving similar improvements in dataset quality.

The key findings of our study include:

\begin{enumerate} \item The effectiveness of our data curation methodology in creating balanced, high-quality datasets that support both complex reasoning and factual knowledge across multiple domains, including astronomy, law, and medicine. \item Significant improvements in model performance on astronomy-related tasks, even with relatively small-scale training data, highlighting the potential for efficient resource utilization. \item The adaptability of our methodology to diverse scientific and professional fields, demonstrating that domain-specific models can outperform general-purpose models in specialized tasks. \end{enumerate}

In conclusion, our work represents a significant step toward more efficient and effective AI tools for specialized scientific and professional domains. As this field continues to evolve, we anticipate that domain-specific language models will play an increasingly important role in supporting research, education, and decision-making across a wide range of disciplines. Moreover, we believe that ongoing collaboration between AI researchers and domain experts will be essential to unlocking the full potential of these models in addressing complex, real-world challenges.

\section*{Acknowledgments}

This research was supported in part by the Delta compute cluster at the National Center for Supercomputing Applications (NCSA). We thank the NCSA for providing computational resources and support that significantly contributed to the success of this work. Furthermore, we acknowledge support by NSF grant AST-2308174

\section{Limitations}

While the ORBIT methodology and the resulting Orbit model show significant promise, it is essential to acknowledge several limitations that may impact their applicability and effectiveness. These limitations are categorized into technical and social aspects to provide a comprehensive understanding of the challenges involved.

\subsection{Technical Limitations}

The primary technical limitations of the ORBIT methodology and the Orbit model are as follows:

\begin{itemize}
    \item \textbf{Domain-Specific Generalizability.} Although ORBIT has proven effective in the field of astronomy, its applicability to other domains remains untested. Domains with less structured data or those that are highly interdisciplinary may require additional adaptations to the filtering and evaluation processes. Defining domain-specific terms and educational value criteria in such fields could pose unique challenges that the current methodology does not address.
    
    \item \textbf{Dependence on Embedding Models.} The embedding-based filtering approach relies heavily on the quality and coverage of pre-trained word embeddings, such as fastText. These embeddings may not fully capture the nuances of highly specialized or emerging astronomical terminology, potentially leading to the exclusion of relevant content or the inclusion of less pertinent material. Enhancing embedding models to better represent domain-specific language could mitigate this limitation.
    
    \item \textbf{Computational and Resource Constraints.} Despite the efficiency gains from using frameworks like DeepSpeed and FlashAttention v2, the fine-tuning process for large models like Orbit demands substantial computational resources. This requirement may limit accessibility for smaller research teams or institutions with limited budgets. Additionally, scaling the methodology to accommodate larger datasets or models with higher parameter counts may encounter practical barriers related to memory and processing power.
    
    \item \textbf{Evaluation Scope.} The current evaluations are primarily focused on astronomy-specific tasks and benchmarks such as MMLU and AstroBench. This narrow scope may limit the generalizability of the findings, as broader benchmarks that include interdisciplinary or collaborative tasks have not been assessed. Expanding the evaluation to encompass a wider range of benchmarks would provide a more comprehensive assessment of the model's utility.
    
    \item \textbf{Dynamic Nature of Scientific Knowledge.} Astronomy is a rapidly evolving field, and the curated dataset represents a specific temporal snapshot. As new discoveries and theories emerge, the model's relevance and accuracy may decline without ongoing updates. Developing methods for efficiently integrating new knowledge into existing models is necessary to maintain their effectiveness over time.
\end{itemize}

Addressing these technical limitations will require future work to explore the adaptability of the ORBIT methodology across domains, enhance embedding models for better domain-specific representation, and develop scalable solutions to manage computational demands.

We acknowledge the assistance of ChatGPT for paraphrasing and shortening text in this document. All content generated with AI was carefully reviewed and validated by the authors.
\section{Ethical Considerations}

The development of domain-specific language models like Orbit raises several ethical considerations that warrant careful examination:

\begin{itemize}
    \item \textbf{Transparency and Open Sourcing.} Open-sourcing the methodology, dataset, and codebase promotes transparency and ensures that other researchers can replicate and validate our findings. However, this accessibility also increases the risk of misuse. For example, malicious actors could adapt the approach to create highly specialized LLMs for unethical purposes, such as generating misleading or pseudoscientific content within specialized domains.
    
    \item \textbf{Mitigation of Misuse.} To mitigate risks of misuse, safeguards such as dataset provenance disclosure, ethical use guidelines, and community oversight should be implemented. Openly documenting the sources and filtering criteria ensures clarity about the data used, while ethical use guidelines can provide clear boundaries for the responsible use of the dataset and methodology. Encouraging the research community to establish and enforce standards for domain-specific LLMs can help prevent misuse.
    
    \item \textbf{Bias and Representation.} While we have curated a dataset with a focus on educational value and scientific rigor, the model could inadvertently propagate biases present in the source data. Historical datasets may reflect outdated or unbalanced perspectives, such as overrepresenting contributions from certain geographic regions or underrepresenting emerging subfields within astronomy. These biases can perpetuate systemic inequities if not carefully addressed.
    
    \item \textbf{Bias Mitigation Strategies.} Post-hoc audits can analyze representation across subfields, geographic regions, and demographics of authorship. Iterative refinement, through periodic dataset updates and expanding coverage of underrepresented areas, can further reduce bias. Engaging a diverse group of domain experts to guide future dataset expansions ensures inclusive curation processes.
    
    \item \textbf{Representation and Inclusivity.} The curated dataset may inadvertently exclude contributions from underrepresented groups or regions, thereby limiting the model's inclusivity. Ensuring diverse representation in the data sources is crucial for developing models that reflect a wide range of perspectives and knowledge bases. Failure to address these disparities can perpetuate existing inequities within the scientific community.
    
    \item \textbf{Transparency and Accountability.} While documenting dataset provenance and filtering criteria promotes transparency, ensuring accountability in the development and deployment of domain-specific models requires ongoing efforts. Establishing clear ethical guidelines and engaging in community oversight are essential steps toward responsible AI development.
\end{itemize}

FineWeb-Edu, our baseline dataset, explicitly addresses the removal of personally identifying and offensive content, as well as trying to address the mentioned issues above. By proactively addressing these ethical considerations, we aim to promote responsible development and deployment of domain-specific language models that support equitable and transparent scientific advancement.

\bibliography{acl_latex}
\appendix

\section{Appendix}
\label{sec:appendix}

\subsection{Evaluation Prompt for Educational Value of Astronomy Texts}
\label{sec:evaluation_prompt}

The following prompt is utilized to assess the educational value of astronomy-related texts. This scoring system assigns a score from 0 to 5 based on the depth, clarity, and relevance of the content. The prompt guides evaluators in determining the quality of information to ensure only high-value educational material is selected for domain-specific training.

\begin{lstlisting}[style=promptStyle]
prompt = f"""Please evaluate the educational value of the following astronomy-related text from a web document. Use this 6-point scoring system:

0 points: No astronomy content at all.
1 point: Minimal astronomy information, or astronomy mixed with non-astronomical content.
2 points: Covers basic astronomical concepts but lacks depth or comprehensive explanation.
3 points: Clear explanation of concepts with relevant examples, educational for a general audience.
4 points: In-depth knowledge, covers advanced concepts or recent discoveries, well-structured and engaging.
5 points: Exceptionally high educational value, expert-level insights, connects multiple concepts, addresses misconceptions, inspires further learning.

Provide a brief justification (up to 100 words) and conclude with the score in the format "Score: X".

Here's the text to evaluate:

{text}"""
\end{lstlisting}

\subsection{Domain-Relevant Astronomy Key Terms}
\label{sec:astronomy_key_terms}

The following list comprises astronomy-related terms used to construct the domain-specific ``astronomy vector'' within the embedding-based filtering process. This selection encompasses key concepts in astrophysics, observational astronomy, and cosmology, ensuring comprehensive coverage of critical terms relevant to domain-specific filtering.

\begin{table}[h]
\centering
\renewcommand{\arraystretch}{1.2} 
\setlength{\tabcolsep}{2pt} 
\begin{tabularx}{\columnwidth}{|X|X|X|}
\hline
\textbf{Terms} & \textbf{Terms} & \textbf{Terms} \\ \hline
Albedo & Aphelion & Apogee \\ \hline
Asteroid & Astronomy & Aurora \\ \hline
Axion & Azimuth & Barycenter \\ \hline
Baryon & Blackbody & Bolide \\ \hline
Brilliance & Cepheid & Comet \\ \hline
Constellation & Corona & Cosmic \\ \hline
Cosmology & DESC & Dyne \\ \hline
Eclipse & Ecliptic & Emission \\ \hline
Erg & Exoplanet & Extinction \\ \hline
Fluence & Frequency & Galaxy \\ \hline
Geocentric & Gibbous & Gravity \\ \hline
Heliocentric & Interferometry & Isotropic \\ \hline
JWST & kpc & Light-Year \\ \hline
LSST & Luminosity & Magnetar \\ \hline
Magnetosphere & Metallicity & Meteor \\ \hline
Meteorite & Microlensing & Moon \\ \hline
Morphology & Multiverse & Nebula \\ \hline
Neutrino & Noctilucent & Nova \\ \hline
Nucleosynthesis & Orbit & Parallax \\ \hline
Parsec & Perihelion & Phase \\ \hline
Photometry & Photosphere & Planck \\ \hline
Planetesimal & Pulsar & Quasar \\ \hline
Quiescence & Recombination & Reddening \\ \hline
Redshift & Reionization & Satellite \\ \hline
Seyfert & Simulation & Singularity \\ \hline
Spectroscopy & SPT & Sublimation \\ \hline
Sunspot & Supercomputer & Supermassive \\ \hline
Supernova & Telescope & Transit \\ \hline
Universe & Voids & Wavelength \\ \hline
Waxing & Wormhole & X-ray \\ \hline
Zenith & Zodiac & Optical \\ \hline
Infrared & Ultraviolet & Microwave \\ \hline
Proton & Neutron & Electron \\ \hline
Flux & Intensity & Companion \\ \hline
Outflow & QSO & Pulse \\ \hline
Progenitor & & \\ \hline
\end{tabularx}
\caption{Domain-Relevant Astronomy Key Terms}
\end{table}

\begin{table}[h]
\centering
\renewcommand{\arraystretch}{1.2} 
\setlength{\tabcolsep}{2pt} 
\begin{tabularx}{\columnwidth}{|X|X|X|}
\hline
\textbf{Terms} & \textbf{Terms} & \textbf{Terms} \\ \hline
Anatomy & Pathology & Physiology \\ \hline
Oncology & Cardiology & Neurology \\ \hline
Radiology & Pharmacology & Surgery \\ \hline
Pediatrics & Dermatology & Gastroenterology \\ \hline
Endocrinology & Hematology & Immunology \\ \hline
Nephrology & Pulmonology & Psychiatry \\ \hline
Rheumatology & Urology & Obstetrics \\ \hline
Gynecology & Orthopedics & Ophthalmology \\ \hline
Otolaryngology & Infectious & Microbiology \\ \hline
Epidemiology & Toxicology & Genetics \\ \hline
Biochemistry & Histology & Embryology \\ \hline
Virology & Bacteriology & Parasitology \\ \hline
Cytology & Prognosis & Diagnosis \\ \hline
Treatment & Therapy & Vaccination \\ \hline
Antibiotic & Antiviral & Pathogen \\ \hline
Tumor & Cancer & Leukemia \\ \hline
Diabetes & Hypertension & Cardiomyopathy \\ \hline
Stroke & Sepsis & Inflammation \\ \hline
Autoimmune & Fibrosis & Circulation \\ \hline
Respiration & Homeostasis & Anesthesia \\ \hline
Trauma & Fracture & Hemorrhage \\ \hline
Venous & Arterial & Renal \\ \hline
Hepatic & Liver & Kidney \\ \hline
Lung & Heart & Brain \\ \hline
Spinal & Nerve & Bone \\ \hline
Muscle & Skin & Blood \\ \hline
Plasma & Lymph & Hormone \\ \hline
Enzyme & Protein & Gene \\ \hline
DNA & RNA & Chromosome \\ \hline
Cell & Tissue & Organ \\ \hline
Organism & Metabolism & Nutrition \\ \hline
Obesity & Malnutrition & Infection \\ \hline
Immunity & Allergy & Vaccine \\ \hline
Mutation & Carcinogen & Biopsy \\ \hline
MRI & CT & X-ray \\ \hline
Ultrasound & PET & Radiotherapy \\ \hline
Chemotherapy & Surgical & Endoscopy \\ \hline
Laparoscopy & Thermography & Pharmacokinetics \\ \hline
Pharmacodynamics & Clinical & Hospital \\ \hline
Ambulance & ICU & Ward \\ \hline
Therapist & Psychologist & Psychiatrist \\ \hline
Physician & Surgeon & Nurse \\ \hline
Paramedic & Dentist & Optometrist \\ \hline
Audiologist & Dietitian & Nutritionist \\ \hline
Emergency & CPR & Defibrillator \\ \hline
Vaccination & Inoculation & Antibody \\ \hline
Antigen & Biomarker & Cytokine \\ \hline
Pathogenesis & Therapeutics & Rehabilitation \\ \hline
Prosthesis & Implant & Transplant \\ \hline
Donor & Recipient & ClinicalTrial \\ \hline
Placebo & DoubleBlind & Epidemic \\ \hline
Pandemic & Outbreak & Quarantine \\ \hline
Contagion & Immunotherapy & PrecisionMedicine \\ \hline
RegenerativeMedicine & Telemedicine & Bioinformatics \\ \hline
Genomics & Proteomics & Metabolomics \\ \hline
Transcriptomics & PersonalizedMedicine & PalliativeCare \\ \hline
Hospice & Prenatal & Postnatal \\ \hline
Neonatal & Geriatrics & Reproductive \\ \hline
Contraception & Fertility & Menopause \\ \hline
Puberty & Hormonal & Chronic \\ \hline
Acute & Degenerative & Congenital \\ \hline
Hereditary & Idiopathic & Nosocomial \\ \hline
Iatrogenic & Symptom & Sign \\ \hline
Prognosis & Complication & Remission \\ \hline
Relapse & Recurrence & SurvivalRate \\ \hline
LifeExpectancy & RiskFactor & Comorbidity \\ \hline
QualityOfLife & Ethics & Consent \\ \hline
Bioethics & Healthcare & Wellness \\ \hline
Preventive & Morbidity & Mortality \\ \hline
Anomaly & Deformity & Lesion \\ \hline
Ulcer & Necrosis & Abscess \\ \hline
Edema & Cyst & Nodule \\ \hline
Polyp & Scar & Adhesion \\ \hline
Prolapse & Hernia & Perforation \\ \hline
Obstruction & Atrophy & Hypertrophy \\ \hline
Hyperplasia & Hypoplasia & Dysplasia \\ \hline
Neoplasia & Metastasis & Differentiation \\ \hline
Invasion & Proliferation & Aneurysm \\ \hline
Thrombosis & Embolism & Ischemia \\ \hline
Infraction & Arrhythmia & Bradycardia \\ \hline
Tachycardia & Fibrillation & Shock \\ \hline
Syncope & Coma & \\ \hline
\end{tabularx}
\caption{Domain-Relevant Medical Key Terms}
\end{table}

\subsection{Training Details}
\label{sec:training_details}

The training of the Orbit-LLaMA model was conducted using the \texttt{DeepSpeed} framework, leveraging \texttt{Zero-2} optimization for efficient memory management and scaling. \texttt{FlashAttention v2} was employed to enhance the efficiency of the self-attention mechanism, improving both memory usage and computational speed.

\textbf{Training Configuration:}
\begin{itemize}
    \item \textbf{Epochs}: 1
    \item \textbf{Block Size}: 512 tokens
    \item \textbf{Effective Batch Size}: 8
    \item \textbf{Learning Rate}: \(2 \times 10^{-5}\)
    \item \textbf{Learning Rate Schedule}: Linear warmup over 500 steps followed by cosine decay
    \item \textbf{Optimizer}: AdamW with parameters \(\beta_1 = 0.9\), \(\beta_2 = 0.95\), and weight decay of 0.01
    \item \textbf{Gradient Clipping}: 1.0
    \item \textbf{Precision}: Mixed precision training enabled with \texttt{bf16} to reduce memory usage and accelerate training
\end{itemize}

\textbf{Optimization Techniques:}
\begin{itemize}
    \item \textbf{DeepSpeed Zero-2 Optimization}: Reduced memory footprint by partitioning optimizer states, gradients, and parameters across GPUs, enabling effective training of large models.
    \item \textbf{FlashAttention v2}: Minimized memory usage during self-attention computations, allowing for faster training without compromising accuracy.
\end{itemize}

\subsection{Qualitative Evaluation Methodology}
\label{sec:qualitative_evaluation}

\subsubsection{Test Questions and Development Process}
A set of 24 test questions was developed by three Ph.D.-track astronomy graduate students and a faculty member from an anonymized university. These questions were designed to evaluate the models’ capabilities across a broad range of topics, including:
\begin{itemize}
    \item \textbf{Basic Definitions and Conceptual Knowledge}: For example, defining astronomical terms.
    \item \textbf{Problem-Solving in Complex or Ambiguous Scenarios}: For instance, addressing under-explored areas of astronomy.
    \item \textbf{Support for Research-Oriented Tasks}: Such as code generation for data analysis or simulations.
\end{itemize}
Each question was carefully reviewed to ensure it was appropriate for benchmarking a wide range of tasks and model competencies.

\subsubsection{Evaluation Framework}
The responses from Orbit, AstroLLaMA, and Meta-LLaMA were evaluated using the following criteria:
\begin{itemize}
    \item \textbf{Accuracy of Content}: How well the response aligned with established scientific understanding.
    \item \textbf{Clarity of Explanation}: The readability and coherence of the generated response.
    \item \textbf{Reasoning}: The logical structure and handling of complex, ambiguous topics.
\end{itemize}

The evaluation process consisted of two phases:
\begin{enumerate}
    \item \textbf{Preference Ratings}: Four graduate students were asked to select their preferred response for each question, considering the aforementioned criteria. This approach allowed reviewers to account for cases where correctness was unclear or when responses were partially correct.
    \item \textbf{Detailed Feedback}: Two astronomy graduate students with Teaching Assistant grading experience analyzed the responses in-depth, identifying strengths, weaknesses, and idiosyncrasies of each model.
\end{enumerate}

\subsubsection{Model-Specific Observations}
Detailed grader comments highlighted the following characteristics:
\begin{itemize}
    \item \textbf{Meta LLaMA-3}: Generated verbose and repetitive answers, occasionally diverging from the original prompt. It was prone to creating irrelevant subtopics and exhibited issues with coherence.
    \item \textbf{Orbit LLaMA}: Produced concise, relevant responses that were easier to follow. It occasionally adopted a teaching-style format, resembling lecture notes or student-generated answers.
    \item \textbf{AstroLLaMA}: Generated longer, more research-like responses. While it included citations, these were not always accurate or relevant. Responses tended to drift off-topic and lacked logical progression.
\end{itemize}

These qualitative observations, combined with the preference ratings, helped identify Orbit LLaMA as the most effective overall. Full responses from all models for each test question, along with grader comments, are provided in the supplementary material.

\subsection{Model Generations Comparison}
\label{sec:model_comparison}

The following section presents a comparison of responses from three models—Meta LLaMA-3, Orbit LLaMA, and AstroLLaMA—to a sample question. Each response is evaluated based on accuracy, clarity, and reasoning.

\subsubsection{Sample Question}
\begin{quote}
What challenges do spacecraft face when crossing the heliopause, and how do these challenges impact the design of interstellar missions?
\end{quote}

\subsubsection{Responses and Evaluations}

\begin{itemize}
    \item \textbf{Meta LLaMA-3 Response:}
    \begin{quote}
    Crossing the heliopause poses several challenges for spacecraft. Firstly, the heliopause region is a highly dynamic environment, with intense solar wind and interstellar magnetic fields. These can cause significant disturbances in the spacecraft's trajectory and navigation, making it difficult to accurately predict the spacecraft's position and velocity. Additionally, the solar wind can cause ionization and plasma flow, which can affect the spacecraft's communication and power systems. The interstellar magnetic field can also induce currents in the spacecraft's structure, which can lead to heating and potential damage. To address these challenges, interstellar missions typically use advanced navigation and control systems, as well as sophisticated shielding and power systems. The shielding is used to protect the spacecraft from ionization and plasma flow, while the power systems are used to generate and distribute electricity to the spacecraft's systems and instruments. Overall, crossing the heliopause is a complex task that requires careful planning and design to ensure the safety and success of the spacecraft.
    \end{quote}
    \textbf{Evaluation:} \textit{Score: 3.} Meta LLaMA-3 provides a reasonable overview of the challenges spacecraft face when crossing the heliopause, mentioning solar wind disturbances, ionization, and the need for advanced systems to mitigate these challenges. However, the response could have provided more specificity about the unique nature of the heliopause and its implications for spacecraft design. There was an emphasis on general navigational and shielding challenges that, while relevant, did not delve deeply into the unique aspects of the heliopause.

    \item \textbf{Orbit LLaMA Response:}
    \begin{quote}
    The heliopause, the boundary between the solar wind and interstellar space, presents unique challenges for spacecraft crossing it. One of the primary challenges is navigation. Without the solar wind to provide a reference point, spacecraft must rely on other methods such as star-tracking or pulsar navigation to determine their position and trajectory. This requires complex navigation systems that can operate in the absence of the familiar solar wind. Another challenge is the radiation environment. The heliopause is surrounded by a region of intense radiation, including cosmic rays and solar wind particles. This radiation can damage spacecraft electronics and pose a threat to the health of the crew. To mitigate these risks, spacecraft must be equipped with advanced radiation shielding and backup systems. Additionally, the distance and communication lag between Earth and the spacecraft make real-time control of the mission impossible. This requires spacecraft to be fully autonomous, capable of making complex decisions and adjustments on their own. Overall, these challenges highlight the need for advanced technology and innovative solutions to enable successful interstellar missions.
    \end{quote}
    \textbf{Evaluation:} \textit{Score: 4.} Orbit LLaMA provides a concise and relevant description of the challenges at the heliopause, such as navigation difficulties due to the lack of solar wind and the increased radiation environment. It clearly explains the need for radiation shielding, autonomy, and advanced navigation systems, directly correlating these challenges with the design implications for interstellar missions. This response was clear, accurate, and focused on the question.

    \item \textbf{AstroLLaMA Response:}
    \begin{quote}
    Crossing the heliopause presents several challenges for interstellar missions, including the interaction of the spacecraft with the solar wind and its magnetic field, as well as the presence of interstellar material such as interstellar dust and the interstellar magnetic field. The heliopause acts as a barrier, preventing the spacecraft from directly interacting with the interstellar medium. However, a boundary layer where interactions between the solar wind and the interstellar medium occur is expected to form at the heliopause. The spacecraft may encounter several layers of plasma and magnetic fields across the boundary layer, which can affect its trajectory and instruments. The spacecraft may also be exposed to interstellar dust, which can be a potential risk to the spacecraft and instruments. The spacecraft designers must consider these challenges and design the spacecraft accordingly, including protective measures and additional instruments to study the interstellar medium.
    \end{quote}
    \textbf{Evaluation:} \textit{Score: 3.} AstroLLaMA offers an extensive explanation of the heliopause and discusses various physical phenomena associated with it. However, while it provides a broad scientific context, it is overly detailed in parts not directly pertinent to the question, which could overwhelm or confuse readers seeking a direct answer. The response lacks specific mention of direct spacecraft design implications, which are crucial for answering the question on interstellar mission design.
\end{itemize}

\section{Supplementary Figures and Tables}
\label{sec:supplementary_figures_tables}

\subsection{Residual Components Distribution}
\label{sec:residual_distribution}

The distribution of residual components for the domain-specific embeddings (\( m = 100 \)) is visualized below. The residuals exhibit a normal distribution centered near zero, validating that noise diminishes with an increasing number of domain-relevant terms. This result supports the robustness of our astronomy vector in representing domain relevance while minimizing noise.

\subsection{Sample Qualitative Evaluation}
\label{sec:sample_qualitative_evaluation}

\subsection{Instructions to Reviewers and Annotator Details}

To evaluate the quality of the model outputs, we recruited four graduate students in astronomy who volunteered to participate in the evaluation process. The primary goal was to compare responses generated by three models—Orbit LLaMA, AstroLLaMA, and Meta LLaMA-3—on a set of astronomy-related questions, focusing on accuracy, clarity, and reasoning.

\subsubsection{Instructions to Annotators}
Annotators were provided with a detailed set of instructions that outlined the evaluation criteria and process. They were asked to:
\begin{enumerate}
    \item \textbf{Read and Assess}: Carefully review the responses generated by the three models for each test question.
    \item \textbf{Evaluate Against Criteria}:
    \begin{itemize}
        \item \textbf{Accuracy}: Determine if the content of the response is factually correct and relevant to the question.
        \item \textbf{Clarity}: Assess whether the response is well-structured, easy to read, and free of ambiguity.
        \item \textbf{Reasoning}: Evaluate the logical structure and whether the response adequately addresses complex or ambiguous aspects of the question.
    \end{itemize}
    \item \textbf{Rank the Responses}: Rank the responses from the three models in order of preference for each question, with justifications for their rankings.
    \item \textbf{Provide Feedback}: Offer detailed comments on strengths and weaknesses for each response, particularly highlighting areas where models excelled or failed.
\end{enumerate}

\subsubsection{Recruitment and Voluntary Participation}
All annotators were graduate students specializing in astronomy. They volunteered for the task, and their contributions were limited to evaluating a manageable set of 24 test questions, ensuring the workload was minimal. No monetary or material compensation was provided as they were willing participants.

\subsubsection{Consent}
All participants were informed about the purpose of the evaluation and the intended use of their feedback for research and publication. Consent was obtained prior to their participation, ensuring ethical compliance.

\subsubsection{Anonymity}
To protect the privacy of the annotators, we do not report individual demographic or geographic characteristics. This decision aligns with their preference for anonymity while contributing to the research.

This structured process ensured high-quality, expert feedback for evaluating model performance while adhering to ethical standards and respecting the privacy of participants.

\subsubsection{Sample Question}
\begin{quote}
What challenges do spacecraft face when crossing the heliopause, and how do these challenges impact the design of interstellar missions?
\end{quote}

\subsubsection{Responses and Evaluations}

\begin{itemize}
    \item \textbf{Meta LLaMA-3 Response:}
    \begin{quote}
    Crossing the heliopause poses several challenges for spacecraft. Firstly, the heliopause region is a highly dynamic environment, with intense solar wind and interstellar magnetic fields. These can cause significant disturbances in the spacecraft's trajectory and navigation, making it difficult to accurately predict the spacecraft's position and velocity. Additionally, the solar wind can cause ionization and plasma flow, which can affect the spacecraft's communication and power systems. The interstellar magnetic field can also induce currents in the spacecraft's structure, which can lead to heating and potential damage. To address these challenges, interstellar missions typically use advanced navigation and control systems, as well as sophisticated shielding and power systems. The shielding is used to protect the spacecraft from ionization and plasma flow, while the power systems are used to generate and distribute electricity to the spacecraft's systems and instruments. Overall, crossing the heliopause is a complex task that requires careful planning and design to ensure the safety and success of the spacecraft.
    \end{quote}
    \textbf{Evaluation:} \textit{Score: 3.} Meta LLaMA-3 provides a reasonable overview of the challenges spacecraft face when crossing the heliopause, mentioning solar wind disturbances, ionization, and the need for advanced systems to mitigate these challenges. However, the response could have provided more specificity about the unique nature of the heliopause and its implications for spacecraft design. There was an emphasis on general navigational and shielding challenges that, while relevant, did not delve deeply into the unique aspects of the heliopause.

    \item \textbf{Orbit LLaMA Response:}
    \begin{quote}
    The heliopause, the boundary between the solar wind and interstellar space, presents unique challenges for spacecraft crossing it. One of the primary challenges is navigation. Without the solar wind to provide a reference point, spacecraft must rely on other methods such as star-tracking or pulsar navigation to determine their position and trajectory. This requires complex navigation systems that can operate in the absence of the familiar solar wind. Another challenge is the radiation environment. The heliopause is surrounded by a region of intense radiation, including cosmic rays and solar wind particles. This radiation can damage spacecraft electronics and pose a threat to the health of the crew. To mitigate these risks, spacecraft must be equipped with advanced radiation shielding and backup systems. Additionally, the distance and communication lag between Earth and the spacecraft make real-time control of the mission impossible. This requires spacecraft to be fully autonomous, capable of making complex decisions and adjustments on their own. Overall, these challenges highlight the need for advanced technology and innovative solutions to enable successful interstellar missions.
    \end{quote}
    \textbf{Evaluation:} \textit{Score: 4.} Orbit LLaMA provides a concise and relevant description of the challenges at the heliopause, such as navigation difficulties due to the lack of solar wind and the increased radiation environment. It clearly explains the need for radiation shielding, autonomy, and advanced navigation systems, directly correlating these challenges with the design implications for interstellar missions. This response was clear, accurate, and focused on the question.

    \item \textbf{AstroLLaMA Response:}
    \begin{quote}
    Crossing the heliopause presents several challenges for interstellar missions, including the interaction of the spacecraft with the solar wind and its magnetic field, as well as the presence of interstellar material such as interstellar dust and the interstellar magnetic field. The heliopause acts as a barrier, preventing the spacecraft from directly interacting with the interstellar medium. However, a boundary layer where interactions between the solar wind and the interstellar medium occur is expected to form at the heliopause. The spacecraft may encounter several layers of plasma and magnetic fields across the boundary layer, which can affect its trajectory and instruments. The spacecraft may also be exposed to interstellar dust, which can be a potential risk to the spacecraft and instruments. The spacecraft designers must consider these challenges and design the spacecraft accordingly, including protective measures and additional instruments to study the interstellar medium.
    \end{quote}
    \textbf{Evaluation:} \textit{Score: 3.} AstroLLaMA offers an extensive explanation of the heliopause and discusses various physical phenomena associated with it. However, while it provides a broad scientific context, it is overly detailed in parts not directly pertinent to the question, which could overwhelm or confuse readers seeking a direct answer. The response lacks specific mention of direct spacecraft design implications, which are crucial for answering the question on interstellar mission design.
\end{itemize}

\section{Additional Mathematical Foundations}
\label{sec:math_foundations_appendix}

\subsection{Efficient Domain-Specific Dataset Curation}
\label{sec:math_foundations_methodology}

The core objective of language models is to estimate the probability distribution over sequences of words by predicting each token based on preceding ones. This is achieved through training on large datasets, where the model minimizes the negative log-likelihood (cross-entropy loss) across the corpus. Model performance tends to improve predictably with the number of parameters, as greater capacity enables capturing more complex patterns—up to a limit governed by dataset quality and complexity \citep{kaplan_scaling_2020}.

Domain-specific models, such as the astronomy-focused variant presented here, face unique challenges in obtaining sufficient, high-quality data, as general-purpose datasets often include noise or irrelevant content. A refined dataset requires filtering methods that prioritize domain relevance without extensive computational costs.

To address this, we developed a method that leverages cosine similarity between token embeddings and a representative aggregated word embedding derived from a predefined list of astronomy-related terms. This approach enables efficient filtering by identifying documents based on their semantic similarity to the target domain.

\subsubsection{Decomposition of Embeddings}
\label{sec:decomposition_embeddings}

We assume that each astronomy-related term's embedding can be decomposed into two components:

\begin{equation}
\mathbf{e}_{t_i} = \mathbf{a} + \mathbf{r}_i,
\end{equation}

where:
\begin{itemize}
    \item \( \mathbf{e}_{t_i} \in \mathbb{R}^d \) is the normalized embedding vector of the \( i \)-th astronomy-related term.
    \item \( \mathbf{a} \in \mathbb{R}^d \) is the domain-specific astronomy component common to all astronomy-related terms.
    \item \( \mathbf{r}_i \in \mathbb{R}^d \) is the random noise component unique to each term, with \( \mathbb{E}[\mathbf{r}_i] = \mathbf{0} \).
\end{itemize}

The astronomy aggregated embedding vector \( \mathbf{A} \) is defined as the average of the embeddings of all astronomy-related terms:

\begin{equation}
\mathbf{A} = \frac{1}{m} \sum_{i=1}^{m} \mathbf{e}_{t_i} = \mathbf{a} + \frac{1}{m} \sum_{i=1}^{m} \mathbf{r}_i.
\end{equation}

By the Law of Large Numbers, as the number of astronomy-related terms \( m \) increases, the average of the random components converges to zero:

\begin{equation}
\lim_{m \to \infty} \frac{1}{m} \sum_{i=1}^{m} \mathbf{r}_i = \mathbf{0}.
\end{equation}

Therefore, for sufficiently large \( m \), the astronomy aggregated vector \( \mathbf{A} \) approximates the domain-specific component \( \mathbf{a} \):

\begin{equation}
\mathbf{A} \approx \mathbf{a}.
\end{equation}

The mean vector \( \mathbf{A} \) also serves as the mathematical minimum point for minimizing the sum of squared Euclidean distances between \( \mathbf{A} \) and each individual astronomy-related embedding \( \mathbf{e}_{t_i} \). Formally, \( \mathbf{A} \) minimizes the following objective:

\begin{equation}
\mathbf{A} = \arg\min_{\mathbf{x} \in \mathbb{R}^d} \sum_{i=1}^{m} \|\mathbf{e}_{t_i} - \mathbf{x}\|^2.
\end{equation}

This property ensures that \( \mathbf{A} \) is the most representative point in the embedding space for the set of astronomy-related terms.

\subsubsection{Error Analysis}
\label{sec:error_analysis}

The error introduced by the random components \( \mathbf{r}_i \) can be quantified by analyzing the difference between the astronomy aggregated vector \( \mathbf{A} \) and the true domain-specific component \( \mathbf{a} \):

\[
\mathbf{E} = \mathbf{A} - \mathbf{a} = \frac{1}{m} \sum_{i=1}^{m} \mathbf{r}_i.
\]

We aim to analyze the expected and actual error rates to ensure that our theoretical results are sound. Specifically, we verify that the random vectors \( \mathbf{r}_i \in \mathbb{R}^d \) are independently and identically distributed (i.i.d.) with mean zero.

To validate the properties of the residual components \( \mathbf{r}_i \), we conducted experiments using the GloVe word embeddings \citep{pennington_glove:_2014}. We assessed whether the residual components for a significant number of astronomy-related terms have an expected value around zero and how the error \( \| \mathbf{E} \| \) behaves as a function of \( m \).

We selected a list of 101 astronomy-related terms (see Section \ref{sec:astronomy_key_terms}) and extracted their corresponding embeddings from the pre-trained GloVe model. We computed the astronomy aggregated vector \( \mathbf{A} \) as the average of \( m \) randomly selected term embeddings and calculated the error vector \( \mathbf{E} = \mathbf{A} - \mathbf{a} \), where \( \mathbf{a} \) is the true average embedding computed using all available astronomy-related terms.

\subsubsection{Computational Efficiency}
\label{sec:computational_efficiency}

To filter a large corpus efficiently, we employ a combination of embedding-based similarity filtering and BERT-based classification. The pipeline's goal is to reduce the dataset to the most relevant documents before applying more computationally expensive processes.

Given a document \( D \) with \( s \) tokens \( \{w_1, w_2, \dots, w_s\} \), each token has a corresponding normalized embedding \( \mathbf{e}_{w_j} \in \mathbb{R}^d \). The document vector \( \mathbf{B} \) is the average of these embeddings:

\[
\mathbf{B} = \frac{\sum_{j=1}^{s} \mathbf{e}_{w_j}}{m}.
\]

The relevance to the astronomy domain is assessed using cosine similarity between \( \mathbf{B} \) and the astronomy vector \( \mathbf{A} \):

\[
\text{Similarity}(D) = \frac{\mathbf{A} \cdot \mathbf{B}}{|A|*|B|}.
\]

A document is retained if this similarity exceeds a threshold \( \tau \).

\subsubsection{Formalized Pipeline}
\label{sec:formalized_pipeline}

\begin{enumerate}
    \item \textbf{Embedding Lookup}: For each token \( w_j \) in document \( D \), retrieve its embedding \( \mathbf{e}_{w_j} \) from a hashmap.
    \hfill \textbf{Runtime}: \( O(1) \)
    
    \item \textbf{Document Vector Computation}: Calculate \( \mathbf{B} = \sum_{j=1}^{s} \mathbf{e}_{w_j} \).
    \hfill \textbf{Runtime}: \( O(s \cdot d) \)
    
    \item \textbf{Similarity Calculation}: Compute cosine similarity between \( \mathbf{A} \) and \( \mathbf{B} \).
    \hfill \textbf{Runtime}: \( O(d) \)
    
    \item \textbf{Thresholding}: Retain the document if the similarity exceeds \( \tau \).
    \hfill \textbf{Runtime}: \( O(1) \)
\end{enumerate}

\noindent
\textbf{Total Complexity per Document}: \( O(s \cdot d) \)

Given \( N \) documents, each with \( s \) tokens on average, the overall complexity for the filtering step is:

\[
O(N \cdot s \cdot d)
\]

\textbf{Optimizations Implemented:}
\begin{itemize}
    \item \textbf{Precomputation of Normalized \( \mathbf{A} \)}: Eliminates repeated division during similarity computation.
    \item \textbf{Vectorized Operations}: Speeds up vector calculations using optimized libraries.
    \item \textbf{Parallel Processing}: Distributes the workload across multiple cores.
\end{itemize}

\section{Mathematical Foundations}
\label{sec:math_foundations_appendix}

\subsection{Efficient Domain-Specific Dataset Curation}
\label{sec:math_foundations_methodology}

The core objective of language models is to estimate the probability distribution over sequences of words by predicting each token based on preceding ones. This is achieved through training on large datasets, where the model minimizes the negative log-likelihood (cross-entropy loss) across the corpus. Model performance tends to improve predictably with the number of parameters, as greater capacity enables capturing more complex patterns—up to a limit governed by dataset quality and complexity \citep{kaplan_scaling_2020}.

Domain-specific models, such as the astronomy-focused variant presented here, face unique challenges in obtaining sufficient, high-quality data, as general-purpose datasets often include noise or irrelevant content. A refined dataset requires filtering methods that prioritize domain relevance without extensive computational costs.

To address this, we developed a method that leverages cosine similarity between token embeddings and a representative aggregated word embedding derived from a predefined list of astronomy-related terms. This approach enables efficient filtering by identifying documents based on their semantic similarity to the target domain.

\subsubsection{Decomposition of Embeddings}
\label{sec:decomposition_embeddings_math}

We assume that each astronomy-related term's embedding can be decomposed into two components:

\begin{equation}
\mathbf{e}_{t_i} = \mathbf{a} + \mathbf{r}_i,
\end{equation}

where:
\begin{itemize}
    \item \( \mathbf{e}_{t_i} \in \mathbb{R}^d \) is the normalized embedding vector of the \( i \)-th astronomy-related term.
    \item \( \mathbf{a} \in \mathbb{R}^d \) is the domain-specific astronomy component common to all astronomy-related terms.
    \item \( \mathbf{r}_i \in \mathbb{R}^d \) is the random noise component unique to each term, with \( \mathbb{E}[\mathbf{r}_i] = \mathbf{0} \).
\end{itemize}

The astronomy aggregated embedding vector \( \mathbf{A} \) is defined as the average of the embeddings of all astronomy-related terms:

\begin{equation}
\mathbf{A} = \frac{1}{m} \sum_{i=1}^{m} \mathbf{e}_{t_i} = \mathbf{a} + \frac{1}{m} \sum_{i=1}^{m} \mathbf{r}_i.
\end{equation}

By the Law of Large Numbers, as the number of astronomy-related terms \( m \) increases, the average of the random components converges to zero:

\begin{equation}
\lim_{m \to \infty} \frac{1}{m} \sum_{i=1}^{m} \mathbf{r}_i = \mathbf{0}.
\end{equation}

Therefore, for sufficiently large \( m \), the astronomy aggregated vector \( \mathbf{A} \) approximates the domain-specific component \( \mathbf{a} \):

\begin{equation}
\mathbf{A} \approx \mathbf{a}.
\end{equation}

The mean vector \( \mathbf{A} \) also serves as the mathematical minimum point for minimizing the sum of squared Euclidean distances between \( \mathbf{A} \) and each individual astronomy-related embedding \( \mathbf{e}_{t_i} \). Formally, \( \mathbf{A} \) minimizes the following objective:

\begin{equation}
\mathbf{A} = \arg\min_{\mathbf{x} \in \mathbb{R}^d} \sum_{i=1}^{m} \|\mathbf{e}_{t_i} - \mathbf{x}\|^2.
\end{equation}

This property ensures that \( \mathbf{A} \) is the most representative point in the embedding space for the set of astronomy-related terms.

\subsubsection{Error Analysis}
\label{sec:error_analysis_math}

The error introduced by the random components \( \mathbf{r}_i \) can be quantified by analyzing the difference between the astronomy aggregated vector \( \mathbf{A} \) and the true domain-specific component \( \mathbf{a} \):

\[
\mathbf{E} = \mathbf{A} - \mathbf{a} = \frac{1}{m} \sum_{i=1}^{m} \mathbf{r}_i.
\]

We aim to analyze the expected and actual error rates to ensure that our theoretical results are sound. Specifically, we verify that the random vectors \( \mathbf{r}_i \in \mathbb{R}^d \) are independently and identically distributed (i.i.d.) with mean zero.

To validate the properties of the residual components \( \mathbf{r}_i \), we conducted experiments using the GloVe word embeddings \citep{pennington_glove:_2014}. We assessed whether the residual components for a significant number of astronomy-related terms have an expected value around zero and how the error \( \| \mathbf{E} \| \) behaves as a function of \( m \).

We selected a list of 101 astronomy-related terms (see Section \ref{sec:astronomy_key_terms}) and extracted their corresponding embeddings from the pre-trained GloVe model. We computed the astronomy aggregated vector \( \mathbf{A} \) as the average of \( m \) randomly selected term embeddings and calculated the error vector \( \mathbf{E} = \mathbf{A} - \mathbf{a} \), where \( \mathbf{a} \) is the true average embedding computed using all available astronomy-related terms.

\begin{figure}[h]
    \centering
    \includegraphics[width=0.4\textwidth]{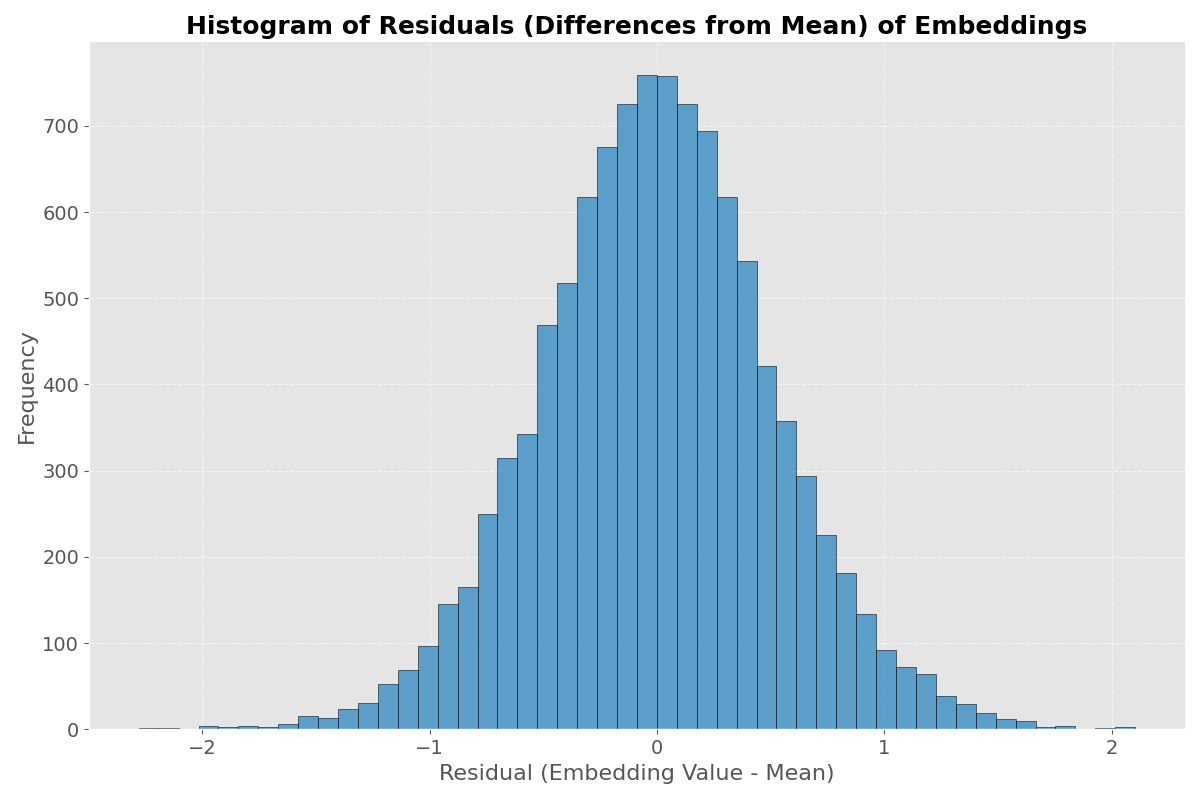}
    \caption{Distribution of residual components for the domain-specific embeddings (\( m = 100 \)). The residuals exhibit a normal distribution centered near zero, validating that noise diminishes with an increasing number of domain-relevant terms. This result supports the robustness of our astronomy vector in representing domain relevance while minimizing noise.}
    \label{fig:histogram_residuals}
\end{figure}

\subsubsection{Computational Efficiency}
\label{sec:computational_efficiency_math}

To filter a large corpus efficiently, we employ a combination of embedding-based similarity filtering and BERT-based classification. The pipeline's goal is to reduce the dataset to the most relevant documents before applying more computationally expensive processes.

Given a document \( D \) with \( s \) tokens \( \{w_1, w_2, \dots, w_s\} \), each token has a corresponding normalized embedding \( \mathbf{e}_{w_j} \in \mathbb{R}^d \). The document vector \( \mathbf{B} \) is the average of these embeddings:

\[
\mathbf{B} = \frac{\sum_{j=1}^{s} \mathbf{e}_{w_j}}{m}.
\]

The relevance to the astronomy domain is assessed using cosine similarity between \( \mathbf{B} \) and the astronomy vector \( \mathbf{A} \):

\[
\text{Similarity}(D) = \frac{\mathbf{A} \cdot \mathbf{B}}{|A| \cdot |B|}.
\]

A document is retained if this similarity exceeds a threshold \( \tau \).

\subsubsection{Formalized Pipeline}
\label{sec:formalized_pipeline_math}

\begin{enumerate}
    \item \textbf{Embedding Lookup}: For each token \( w_j \) in document \( D \), retrieve its embedding \( \mathbf{e}_{w_j} \) from a hashmap.
    \hfill \textbf{Runtime}: \( O(1) \)
    
    \item \textbf{Document Vector Computation}: Calculate \( \mathbf{B} = \sum_{j=1}^{s} \mathbf{e}_{w_j} \).
    \hfill \textbf{Runtime}: \( O(s \cdot d) \)
    
    \item \textbf{Similarity Calculation}: Compute cosine similarity between \( \mathbf{A} \) and \( \mathbf{B} \).
    \hfill \textbf{Runtime}: \( O(d) \)
    
    \item \textbf{Thresholding}: Retain the document if the similarity exceeds \( \tau \).
    \hfill \textbf{Runtime}: \( O(1) \)
\end{enumerate}

\noindent
\textbf{Total Complexity per Document}: \( O(s \cdot d) \)

Given \( N \) documents, each with \( s \) tokens on average, the overall complexity for the filtering step is:

\[
O(N \cdot s \cdot d)
\]

\textbf{Optimizations Implemented:}
\begin{itemize}
    \item \textbf{Precomputation of Normalized \( \mathbf{A} \)}: Eliminates repeated division during similarity computation.
    \item \textbf{Vectorized Operations}: Speeds up vector calculations using optimized libraries.
    \item \textbf{Parallel Processing}: Distributes the workload across multiple cores.
\end{itemize}

\end{document}